# ChemCLIP: Bridging Organic and Inorganic Anticancer Compounds Through Contrastive Learning

Mohamad Koohi-Moghadam[1,*], Hongzhe Sun[2], Hongyan Li[2], Kyongtae Tyler Bae[1]

[1] Department of Diagnostic Radiology, Li Ka Shing Faculty of Medicine, The University of Hong Kong, Pok Fu Lam Road, Hong Kong SAR, PR China

[2] Department of Chemistry, The University of Hong Kong, Pok Fu Lam Road, Hong Kong S.A.R., PRC

[*] Corresponding author: M.K (koohi@hku.hk)

## Abstract

The discovery of anticancer therapeutics has traditionally treated organic small molecules and metal-based coordination complexes as separate chemical domains, limiting knowledge transfer despite their shared biological objectives. This disparity is particularly pronounced in available data, with extensive screening databases for organic compounds compared to only a few thousand characterized metal complexes. Here, we introduce ChemCLIP, a dual-encoder contrastive learning framework that bridges this organic-inorganic divide by learning unified representations based on shared anticancer activities rather than structural similarity. We compiled complementary datasets comprising 44,854 unique organic compounds and 5,164 unique metal complexes, standardized across 60 cancer cell lines. By training parallel encoders with activity-aware hard negative mining, we mapped structurally distinct compounds into a shared 256-dimensional embedding space where biologically similar compounds cluster together regardless of chemical class. We systematically evaluated four molecular encoding strategies—Morgan fingerprints, ChemBERTa, MolFormer, and Chemprop—through quantitative alignment metrics, embedding visualizations, and downstream classification tasks. Morgan fingerprints achieved superior performance with an average alignment ratio of 0.899 and downstream classification AUCs of 0.859 (inorganic) and 0.817 (organic). The high classification performance using frozen embeddings—without task-specific fine-tuning—demonstrates that biologically meaningful features are encoded in the learned representations, enabling practical applications in computational drug screening. This work establishes contrastive learning as an effective strategy for unifying disparate chemical domains and provides empirical guidance for encoder selection in multi-modal chemistry applications, with implications extending beyond anticancer drug discovery to any scenario requiring cross-domain chemical knowledge transfer.

## Introduction

The development of anticancer therapeutics has traditionally progressed along two parallel but largely independent paths: organic small molecules and metal-based coordination complexes [1, 2]. While organic compounds dominate the pharmaceutical landscape due to decades of systematic screening and optimization, metal-based drugs—exemplified by the clinical success of platinum-based chemotherapeutics—offer distinct mechanisms of action and potential advantages in addressing drug resistance [3, 4]. However, the vast disparity in available data between these domains presents a significant challenge for computational drug discovery. The

National Cancer Institute's NCI60 screening program has generated activity data for over 44,000 organic compounds [5], whereas comprehensive databases of metal complexes contain only a few thousand compounds. This data imbalance, combined with the fundamental structural differences between organic molecules and coordination complexes, has traditionally necessitated separate computational models and limited knowledge transfer between organic and inorganic domains [6-10].

Recent advances in contrastive learning—most notably demonstrated by CLIP (Contrastive Language-Image Pre-training) [11] in the vision-language domain—suggest a promising approach to bridging such disparate modalities [12]. By learning to align representations based on shared functional properties rather than structural similarity, contrastive learning frameworks can discover meaningful correspondences between fundamentally different data types. In the context of anticancer drug discovery, this paradigm shift offers an opportunity to create unified computational models that leverage the extensive knowledge embedded in organic compound databases to inform predictions about metal-based therapeutics, while simultaneously enabling direct comparison and ranking of structurally diverse candidates based on their biological activities.

This study introduces ChemCLIP, a dual-encoder contrastive learning framework specifically designed to bridge the organic-inorganic divide in anticancer compound space. We compiled complementary datasets comprising 44,854 unique organic compounds from the NCI60 [5] screening program and 5,164 unique metal complexes from MetalCytoToxDB [13], standardized across 60 shared cancer cell lines. By training parallel encoders to map these structurally distinct compound classes into a unified 256-dimensional embedding space, we hypothesized that compounds exhibiting similar biological activities would cluster together regardless of their chemical class. To rigorously evaluate this approach, we compared four molecular encoding strategies—Morgan fingerprints [14], ChemBERTa [15], MolFormer [16], and Chemprop [17]—assessing their ability to learn activity-relevant representations through quantitative alignment metrics, embedding space visualizations, and downstream classification performance.

Our work addresses three fundamental questions: (1) Can contrastive learning successfully align organic and inorganic chemical spaces based on shared biological activities? (2) Which molecular encoding strategies best support cross-domain generalization in the context of metal complexes and organic molecules? (3) Does the learned embedding space provide effective representations for downstream tasks such as compound activity classification? The answers to these questions have direct implications for computational drug screening pipelines and broader applications of multi-modal learning in chemistry, where bridging diverse chemical domains could accelerate discovery across multiple therapeutic areas.

**Method**

**Dataset Preparation**

We compiled two complementary datasets to enable contrastive learning between organic and inorganic anticancer compounds. The organic compound dataset was derived from the National

Cancer Institute's NCI60 Human Tumor Cell Lines Screen [5], a publicly available database containing dose-response data for thousands of small molecules tested against 60 human cancer cell lines representing nine tissue types. The inorganic compound dataset was obtained from MetalCytoToxDB, a specialized database of metal-based anticancer compounds across various cancer cell lines.

*For Organic Dataset Processing.* Compounds were classified as active or inactive based on their mean growth inhibition percentage across tested concentrations (threshold < 50). Lower growth inhibition percentage values indicate stronger growth inhibitory activity, making this threshold appropriate for identifying compounds with meaningful anticancer effects. SMILES representations were retrieved from PubChem using NSC (National Service Center) identifiers. Compounds lacking valid SMILES structures were excluded from further analysis. Metal-containing compounds identified in the NCI60 database were extracted and transferred to the inorganic dataset to ensure clear separation between organic and inorganic chemical spaces. To ensure compatibility between organic and inorganic datasets, we filtered the NCI60 data to retain only the 60 cell lines that were present in both datasets. This standardization was essential for enabling meaningful contrastive learning across chemical classes.

*For Inorganic Dataset Processing.* Cell line names in MetalCytoToxDB [13] were standardized to match NCI60 nomenclature. We created a systematic mapping between database-specific cell line identifiers and standard NCI60 cell line names. Entries without valid NCI60 cell line mappings were removed to maintain dataset consistency. Metal complexes were classified based on their IC50 and compounds with IC50 < 10 μM were labeled as active. For each metal complex, we extracted chemical features including metal type (one-hot encoded for Ru, Ir, Re, Os, Rh, Cu, Pt, Au, Co, and Ti), oxidation state, atomic number, and valence electron count. These features enabled the model to learn metal-specific pharmacological patterns.

**Dataset Statistics**

*Inorganic dataset.* The processed inorganic dataset comprised 13,656 records representing 5,164 unique metal complexes tested across 60 cell lines. The activity distribution showed 3,165 active compounds (23.18%) and 10,491 inactive compounds (76.82%), yielding an active-to-inactive ratio of 1:3.3. The dataset was dominated by ruthenium complexes (52.05%), followed by titanium (17.57%), and iridium (9.89%), with ten transition metals represented in total. The 60 cell lines covered nine major cancer types, with substantial representation from colon cancer, melanoma, ovarian cancer, and leukemia. Notably, 60.65% of records originated from MetalCytoToxDB, while 39.35% came from metal compounds screened in the NCI60 program (Table S1-3).

*Organic Dataset.* The processed organic dataset contained 1,812,339 records representing 44,854 unique organic compounds tested across 71 cell lines. The activity distribution was more imbalanced than the inorganic dataset, with 44,835 active compounds (2.47%) and 1,767,504 inactive compounds (97.53%), resulting in an active-to-inactive ratio of 1:39.4. The dataset showed broad coverage across cancer types, with colon cancer (17.37%), melanoma (16.56%), and ovarian cancer (14.16%) being the most represented. On average, each unique compound was tested against 40.4 cell lines, demonstrating extensive cross-panel screening that enables robust activity pattern learning. The dataset's compound diversity, with an average of approximately 631 compounds tested per cell line, provided comprehensive chemical space coverage for training the contrastive learning model Table S4, 5.

**Model Architecture and Training Methodology**

ChemCLIP implements a dual-encoder contrastive learning architecture adapted from CLIP (Contrastive Language-Image Pre-training) for bridging inorganic and organic chemical compound spaces. The model learns a shared embedding space where compounds exhibiting similar biological activities across cancer cell lines achieve high similarity regardless of their chemical class (organic versus inorganic). This approach enables cross-domain retrieval, knowledge transfer from the larger organic compound database to metal-based drugs, and downstream activity prediction tasks.

*Dual-Encoder Architecture.* The model comprises two parallel encoding branches that process inorganic metal complexes and organic compounds separately before projecting them into a unified 256-dimensional embedding space. The inorganic branch incorporates specialized metal feature integration, while the organic branch processes standard molecular representations. Both branches employ identical projection architectures to ensure symmetry in the learned embedding space.

We evaluated four distinct molecular encoding strategies to identify the optimal representation for our contrastive learning framework. Morgan fingerprints (radius=2, 2048 bits) provided a cheminformatics baseline with fast computation but fixed representations. ChemBERTa and MolFormer represented transformer-based approaches pre-trained on large chemical databases. For graph-based molecular encoding, we implemented Chemprop's Directed Message Passing Neural Network (D-MPNN) [18].

The inorganic encoder enhanced base molecular representations with metal-specific information critical for understanding metal complex pharmacology. For each metal complex, the ligand SMILES was processed through the base encoder, while metal features were separately encoded. Metal type was represented through one-hot encoding across 10 transition metals (Au, Co, Cu, Ir, Os, Pt, Re, Rh, Ru, Ti), supplemented by scalar features for oxidation state and atomic number. This architecture enabled the model to learn metal-specific pharmacological patterns while maintaining compatibility with organic compound representations.

*Contrastive Learning with Hard Negative Mining.* The model training employed a two-component loss function that combined standard contrastive learning with activity-aware hard negative mining. The first component was the bidirectional InfoNCE contrastive loss [19]. Within each training batch, we computed a similarity matrix between all inorganic and organic compound embeddings using normalized dot products scaled by a temperature parameter ($\tau = 0.07$). Each inorganic compound was paired with exactly one organic compound from the same cell line, forming the positive pairs represented by diagonal elements in the similarity matrix. All other inorganic-organic combinations within the batch served as negative samples. The loss function encouraged high similarity between positive pairs while simultaneously pushing apart negative pairs through bidirectional cross-entropy computed in both inorganic-to-organic and organic-to-inorganic directions.

The second component introduced activity-aware hard negative mining through triplet margin loss [20]. Unlike the random negatives in the contrastive loss, hard negatives were specifically selected based on activity labels to create more challenging learning scenarios. For each active inorganic compound, we selected an active organic compound from the same cell line as the positive example and an inactive organic compound from the same cell line as the hard negative. This pairing strategy forced the model to distinguish between active and inactive compounds within the same biological context, rather than relying on easier distinctions between compounds from different cell lines. The triplet loss enforced that the similarity between an inorganic compound and its active organic match should exceed the similarity to its inactive organic counterpart. The final training objective summed the contrastive loss and triplet loss with equal weighting, enabling the model to learn both broad chemical similarity patterns and fine-grained activity-specific discriminations.

***Training Strategy and Data Management.*** To address the severe class imbalance in the organic dataset (2.47% active versus 97.53% inactive), we retained all active compounds while subsampling inactive compounds at a 5:1 ratio relative to active compounds. We implemented compound-based data splitting to ensure rigorous evaluation of generalization to unseen chemical structures. Unique compounds from both datasets were randomly partitioned into 70% training, 15% validation, and 15% test sets. All compound-cell line pairs were assigned to their respective compound's split, preventing data leakage through compound replication across cell lines. This strategy provided stricter evaluation of model performance on completely novel chemical structures compared to random splitting of compound-cell line pairs. Model training employed the AdamW optimizer with learning rate $1\times10^{-3}$, default weight decay (0.01), and beta parameters (0.9, 0.999). Training proceeded for 100 epochs with batch size 128, gradient clipping at maximum norm 1.0, and dropout rate 0.1. This configuration balanced training stability with efficient convergence across all encoder architectures evaluated.

## Results

### Embedding Space Visualization

To evaluate the quality of learned embeddings and assess the effectiveness of different molecular encoders, we performed t-SNE (t-distributed Stochastic Neighbor Embedding) [21] visualization of compound representations both before and after contrastive training. The visualizations compared four molecular encoding strategies—Morganfingerprints, ChemBERTa, MolFormer, and Chemprop—across two conditions: initial frozen encoders, and trained models projecting both compound types into the shared 256-dimensional embedding space (Figure 1).

The four encoders exhibited markedly different capabilities in learning task-relevant representations through contrastive training. Morgan fingerprints demonstrated superior embedding quality, achieving clear separation between active and inactive compounds within both inorganic and organic chemical spaces after training. This separation was evident across both compound types, indicating that the model successfully learned activity-relevant patterns that generalized across the organic-inorganic divide. ChemBERTa and MolFormer, both transformer-based architectures pre-trained on large chemical databases, showed intermediate performance. While these encoders achieved reasonable separation between active and inactive compounds after training, the clustering was less pronounced compared to Morganfingerprints. Chemprop's performance revealed a significant limitation in the contrastive learning paradigm. The trained Chemprop embeddings primarily separated inorganic compounds from organic compounds but failed to achieve meaningful separation between active and inactive compounds within each chemical class. This suggests that the graph-based message-passing architecture, while effective for capturing molecular structure, struggled to learn fine-grained activity distinctions under the dual-encoder contrastive framework.

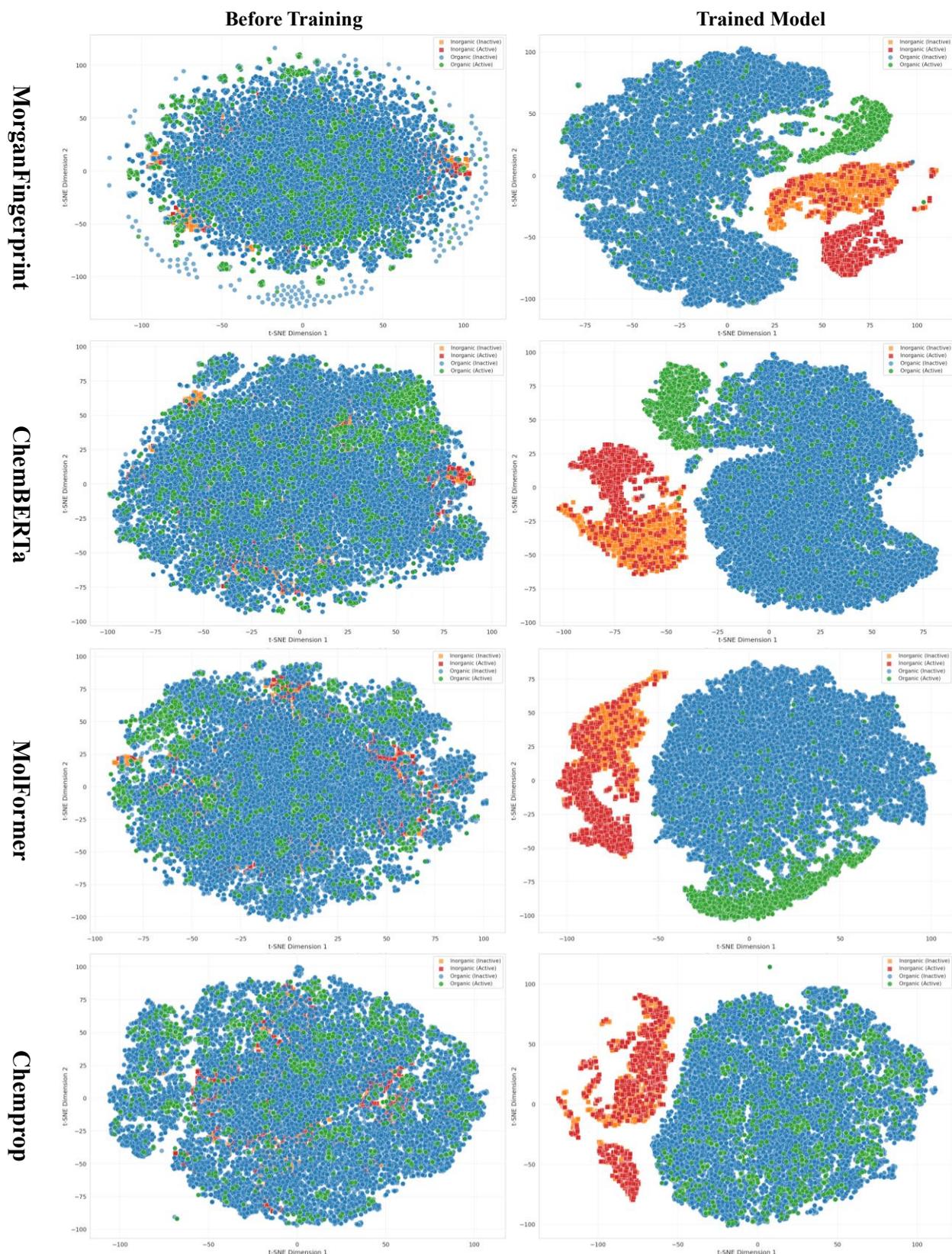

**Figure 1.** Comparison of four molecular encoders. Left panels show encoder embeddings before training and right panels show trained ChemCLIP embeddings with joint t-SNE in the shared 256-dimensional space. Colors indicate compound type (inorganic: red squares, organic: blue circles) and activity (darker: active, lighter: inactive). MorganFingerprint shows the clearest separation between active and inactive compounds in both types.

**Quantitative Assessment of Cross-Modal Alignment**

To quantitatively assess the quality of cross-modal alignment in the learned embedding space, we performed a comprehensive statistical analysis using cluster center distances. This analysis directly tests the central hypothesis of ChemCLIP: compounds with similar biological activities should be closer in the shared embedding space, regardless of whether they are inorganic or organic. We computed centroids for four distinct groups—Inorganic Active (IA), Inorganic Inactive (II), Organic Active (OA), and Organic Inactive (OI)—and defined two complementary metrics:

**Alignment Ratio** (lower is better) measures whether same-activity compounds from different modalities are closer than different-activity compounds:

$$\text{Alignment Ratio} = \frac{1}{2}\left[\frac{d(\text{IA,OA})}{d(\text{IA,OI})} + \frac{d(\text{II,OI})}{d(\text{II,OA})}\right]$$

Values below 1.0 indicate successful alignment.

**Cross-Modal Separation Ratio** (higher is better) quantifies the overall separation between different-activity and same-activity pairs:

$$\text{Separation Ratio} = \frac{\text{Avg}[d(\text{IA,OI}), d(\text{II,OA})]}{\text{Avg}[d(\text{IA,OA}), d(\text{II,OI})]}$$

Values above 1.0 indicate meaningful activity-based clustering.

Figure 2 presents the alignment and separation ratios for all four encoders. MorganFingerprint achieved the best performance (average alignment ratio: 0.899, separation ratio: 1.127), followed by ChemBERTa (0.903, 1.119) and MolFormer (0.920, 1.093). All three encoders exceeded the threshold of 1.0, confirming successful cross-modal alignment. In contrast, Chemprop showed ratios of exactly 1.000, indicating complete failure to organize the embedding space by activity.

The cross-modal distance matrix reveals that for MorganFingerprint, inorganic active compounds are 21.3% closer to organic active compounds (d = 1.147) than to organic inactive compounds (d = 1.457), yielding an active alignment ratio of 0.787. ChemBERTa exhibited similar behavior (0.796), while MolFormer showed weaker alignment (0.837). The uniform distances (~1.265) for Chemprop across all cluster pairs indicate embedding space collapse, where only compound type is preserved without activity discrimination. The combined performance score integrates both metrics: MorganFingerprint (0.228) > ChemBERTa (0.216) > MolFormer (0.174) > Chemprop (0.000). While differences between MorganFingerprint and ChemBERTa are modest (5.6%), MorganFingerprint consistently outperforms across all metrics.

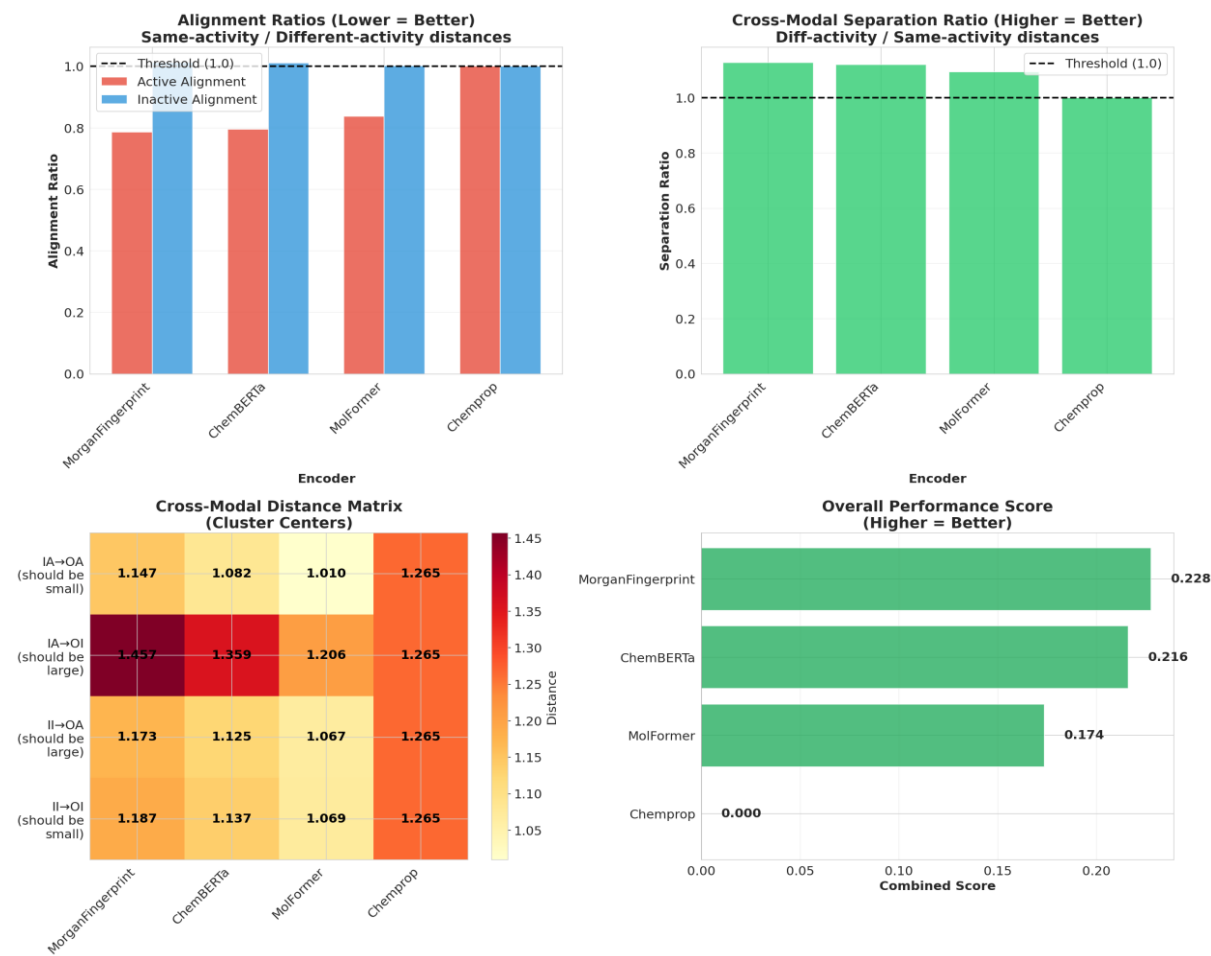

**Figure 2: Quantitative Comparison of Cross-Modal Alignment Analysis.** (A) Alignment ratios for all encoders; lower values indicate better alignment of same-activity compounds. (B) Cross-modal separation ratios; higher values indicate better separation of different-activity compounds. (C) Distance heatmap showing all pairwise cluster center distances. (D) Combined performance scores. MorganFingerprint consistently outperforms other encoders. Abbreviations: IA, Inorganic Active; II, Inorganic Inactive; OA, Organic Active; OI, Organic Inactive.

**Downstream Classification Performance**

To evaluate the discriminative power and practical utility of the learned embedding space, we trained binary classifiers on the frozen ChemCLIP embeddings to predict compound activity (active vs. inactive). This downstream task directly tests whether the 256-dimensional shared embedding space captures biologically relevant features that enable accurate activity prediction without further encoder fine-tuning. We trained separate classifiers for inorganic and organic compounds using the same test split employed during contrastive learning, ensuring that test set compounds were completely unseen during both contrastive training and classification training.

*Classification Methodology.* For each encoder, we trained two independent binary classifiers—one for inorganic compounds and one for organic compounds—using a simple three-layer multilayer perceptron architecture [22]. Classifiers were trained using the same compound-based splits employed during contrastive learning, ensuring test set compounds remained completely unseen during both training phases. Encoder weights were frozen throughout classifier training to evaluate embedding quality independently of task-specific adaptation.

The datasets exhibit severe class imbalance, with only 2.5% of organic compounds and 23.2% of inorganic compounds labeled as active. To address this challenge, we applied weighted binary cross-entropy loss where positive class weights equal the ratio of negative to positive samples, and optimized classification thresholds on the validation set using F1 score rather than default probability cutoffs. We employed AUC-ROC [23] as the primary evaluation metric due to its robustness to class imbalance, complemented by F1 score to assess precision-recall balance [24].

**Table 1: Overall Classification Performance.** Bold = best in column; AUC = Area Under ROC Curve; F1 = F1 Score; Acc = Accuracy

| Encoder | Inorganic AUC | Inorganic F1 | Inorganic Acc | Organic AUC | Organic F1 | Organic Acc | Avg. AUC |
|---|---|---|---|---|---|---|---|
| **MorganFingerprint** | **0.859** | **0.720** | **0.841** | 0.817 | 0.621 | 0.846 | **0.838** |
| ChemBERTa | 0.834 | 0.669 | 0.811 | 0.780 | 0.623 | 0.842 | 0.807 |
| MolFormer | 0.729 | 0.497 | 0.717 | **0.874** | **0.732** | **0.852** | 0.801 |
| Chemprop | 0.600 | 0.493 | 0.444 | 0.501 | 0.417 | 0.264 | 0.550 |

*Classification Results and Analysis.* Table 1 presents comprehensive classification performance across all four encoders. MorganFingerprint achieved the best overall performance with the highest average AUC of 0.838, demonstrating superior discriminative power across both compound types. The encoder particularly excelled on inorganic compounds, attaining 0.859 AUC and 0.720 F1 score while maintaining competitive organic performance at 0.817 AUC. Detailed analysis (Supplementary Tables S6 and S7) reveals that MorganFingerprint embeddings achieve exceptional precision of 0.884 on organic compounds—a critical advantage for drug screening applications where false positives incur substantial experimental costs.

MolFormer exhibited domain-specific excellence, achieving the best organic performance across all encoders with 0.874 AUC and 0.732 F1 score. However, this strength came at the cost of poor inorganic classification, with AUC dropping to 0.729. This 14.5-point gap between organic and inorganic performance reveals strong distributional bias from pre-training on 1.1 billion predominantly organic molecules from PubChem. While large-scale pre-training provides powerful inductive biases for organic chemistry, these advantages do not transfer to coordination complexes with distinct electronic structures and bonding patterns not well-represented in pre-training corpora. ChemBERTa provided balanced and reliable performance across both compound types, achieving 0.834 AUC on inorganic and 0.780 AUC on organic compounds with only 5.4% variation between domains. This consistency demonstrates robust cross-domain generalization without domain-specific overfitting. However, ChemBERTa's peak performance remained lower than specialized encoders, suggesting a tradeoff between generalization breadth and task-specific optimization. Chemprop exhibited complete classification failure, with AUC scores barely above random chance at 0.600 for inorganic and 0.501 for organic compounds. Analysis of detailed metrics reveals a pathological failure mode: the classifier defaults to predicting all compounds as active, achieving perfect recall but near-zero precision. This behavior confirms the embedding space collapse previously observed in t-SNE visualizations and statistical alignment metrics.

**Discussion**

**Principal Findings and Contributions**

This study introduces ChemCLIP, a contrastive learning framework that bridges the traditionally separate domains of organic and inorganic anticancer compounds [25] through a unified embedding space. Our results demonstrate that cross-modal alignment between metal complexes and organic molecules can successfully capture shared biological activity patterns, enabling knowledge transfer from the extensively studied organic chemical space to the comparatively under-explored inorganic domain. Among four evaluated molecular encoders, MorganFingerprint emerged as the optimal choice, achieving superior performance across embedding quality (average alignment ratio: 0.899), visualization clarity, and downstream classification tasks (average AUC: 0.838). The strong classification performance using frozen embeddings—without task-specific fine-tuning—validates that biologically relevant features are encoded in the learned representations, offering practical advantages for drug discovery applications where labeled training data may be limited.

**Activity-Aware Hard Negative Mining and Its Role in Learning Discrimination**

The incorporation of activity-aware hard negative mining through triplet margin loss proved essential for achieving the observed classification performance. Standard contrastive learning with random negative sampling encourages embeddings to separate compounds from different cell lines, which may correlate with activity differences but does not explicitly enforce activity-based discrimination. By specifically pairing each active inorganic compound with an active organic compound (positive) and an inactive organic compound (hard negative) from the same cell line, the triplet loss forces the model to learn subtle distinctions between active and inactive compounds within identical biological contexts.

This design choice addresses a fundamental challenge in multi-label biological activity prediction: compounds may exhibit cell line-specific effects, and simple cross-entropy losses can exploit cell line identity as a shortcut rather than learning generalizable activity patterns. The hard negative mining strategy eliminates this shortcut by ensuring that positive and negative pairs share the same cell line, compelling the model to discover chemical features that predict activity independent of the testing context. The clear separation between active and inactive compounds in the t-SNE visualizations, particularly for MorganFingerprint, validates that this approach successfully guides the contrastive learning process toward biologically meaningful representations.

The equal weighting of contrastive and triplet losses represents a balance between two complementary objectives: the bidirectional InfoNCE loss establishes broad cross-modal alignment, while the triplet loss refines this alignment to distinguish active from inactive compounds. Alternative weighting schemes or curriculum learning approaches—where the relative importance of each loss component varies during training—might further improve embedding quality, particularly in the early training phases where establishing basic cross-modal correspondence may be more important than fine-grained activity discrimination.

**Implications for Drug Discovery and Screening**

The strong classification performance achieved using frozen ChemCLIP embeddings (0.80-0.87 AUC for successful encoders) demonstrates practical utility for computational drug screening pipelines. Traditional approaches to anticancer drug discovery often treat organic and inorganic compounds as separate chemical spaces, requiring independent computational models, screening protocols, and optimization strategies. ChemCLIP's unified framework enables direct comparison and ranking of structurally diverse compounds based on predicted biological activity, potentially accelerating the identification of metal-based alternatives to organic drugs or revealing opportunities for hybrid therapeutic strategies.

The high precision achieved by MorganFingerprint on both organic (0.884) and inorganic (0.723) compounds proves particularly valuable for prioritizing compounds for experimental validation. In high-throughput screening campaigns where experimental capacity limits the number of compounds that can be tested, ranking compounds by predicted activity and selecting top candidates can dramatically reduce costs while maintaining high hit rates. A precision of 0.88 implies that among compounds predicted as active, 88% truly exhibit anticancer activity—a hit rate far exceeding typical screening success rates and justifying the computational investment required for embedding-based prioritization.

The transfer learning capability demonstrated by training simple classifiers on frozen embeddings offers additional advantages for specialized applications. Researchers investigating specific cancer subtypes, resistance mechanisms, or combination therapies can fine-tune lightweight classifiers on small datasets without retraining the computationally expensive encoder. This approach proves especially valuable for inorganic compounds, where limited experimental data has historically constrained the application of machine learning methods. By leveraging the knowledge encoded in the organic compound training set through the shared embedding space, ChemCLIP enables effective activity prediction for metal complexes even when direct training examples are scarce.

**Broader Implications for Chemical Representation Learning**

Beyond the specific application to anticancer compounds, this work illustrates general principles for bridging chemical domains through contrastive learning. The success of domain-agnostic structural encodings suggests that multi-domain chemical applications—such as predicting environmental toxicity across organic pollutants and inorganic heavy metals [26], or optimizing catalytic activity across homogeneous and heterogeneous catalysts—would benefit from similar approaches. The failure of pre-trained transformers to achieve balanced cross-domain performance highlights the importance of matching encoder architectures to task requirements rather than defaulting to the largest available foundation models.

The ChemCLIP framework could extend naturally to other multi-modal chemistry problems where explicit similarity labels are unavailable but functional equivalence can be inferred from shared properties. Contrastive learning between molecules and their spectroscopic signatures (NMR, IR, mass spectra) could accelerate structure elucidation for unknown compounds. Alignment between chemical structures and natural language descriptions of their properties could enable text-based chemical search and improve human-AI interfaces for chemistry applications. The core principle—learning shared embeddings by enforcing similarity between functionally equivalent but structurally dissimilar entities—applies broadly across scientific domains wherever multiple representations of the same underlying phenomena exist.

**Conclusion**

This study demonstrates that contrastive learning can successfully bridge organic and inorganic chemical spaces, creating unified representations that capture shared biological activities despite fundamental structural differences. MorganFingerprint's domain-agnostic structural encoding emerges as the optimal choice for this task, achieving superior cross-modal alignment, embedding quality, and downstream classification performance. The strong results obtained with simple classifiers trained on frozen embeddings validate the practical utility of this approach for drug discovery applications, enabling knowledge transfer from extensively studied organic compounds to comparatively under-explored metal-based drugs. These findings establish contrastive learning as a promising strategy for multi-domain chemistry applications and provide guidance for encoder selection in scenarios requiring balanced performance across structurally diverse chemical classes.

**Conflict of interest**
The authors of this study declare that they do not have any conflict of interest.

**Data availability statement**
The data we used are publicly available and have already been referenced in our manuscript. The training and inference code is available at https://github.com/.